\documentclass[manuscript,screen]{acmart}

\AtBeginDocument{%
  \providecommand\BibTeX{{%
    \normalfont B\kern-0.5em{\scshape i\kern-0.25em b}\kern-0.8em\TeX}}}

\setcopyright{acmlicensed}
\copyrightyear{2018}
\acmYear{2018}
\acmDOI{XXXXXXX.XXXXXXX}

\usepackage{longtable}

\begin{document}

\title{Prompting and Fine-Tuning Open-Sourced Large Language Models for Stance Classification}

\author{Iain J. Cruickshank}
\affiliation{%
  \institution{Army Cyber Institute}
  \streetaddress{2101 New South Post Road}
  \city{Highland Falls}
  \state{New York} 
  \country{USA}
  \postcode{10996}}
\email{iain.cruickshank@westpoint.edu}

\author{Lynnette Hui Xian Ng}
\affiliation{%
  \institution{Carnegie Mellon University}
  \streetaddress{5000 Forbes Ave}
  \city{Pittsburgh}
  \state{Pennslyvania}
  \country{USA}
  \postcode{15213}}
\email{lynnetteng@cmu.edu}

\renewcommand{\shortauthors}{Cruickshank et al}

\begin{abstract}
Stance classification, the task of predicting the viewpoint of an author on a subject of interest, has long been a focal point of research in domains ranging from social science to machine learning. Current stance detection methods rely predominantly on manual annotation of sentences, followed by training a supervised machine learning model. However, this manual annotation process requires laborious annotation effort, and thus hampers its potential to generalize across different contexts. In this work, we investigate the use of Large Language Models (LLMs) as a stance detection methodology that can reduce or even eliminate the need for manual annotations. We investigate 10 open-source models and 7 prompting schemes, finding that LLMs are competitive with in-domain supervised models but are not necessarily consistent in their performance. We also fine-tuned the LLMs, but discovered that fine-tuning process does not necessarily lead to better performance. In general, we discover that LLMs do not routinely outperform their smaller supervised machine learning models, and thus call for stance detection to be a benchmark for which LLMs also optimize for.
The code used in this study is available at \url{https://github.com/ijcruic/LLM-Stance-Labeling}.

\end{abstract}




\maketitle

\section{Introduction}
Stance classification, also known as stance detection, is a natural language processing task of eliciting the author's viewpoint towards a subject of interest. That is, understanding the stance towards a target. Approaches towards this task typically use a supervised machine learning model to learn stance labels (i.e., agree, disagree, neutral) that correspond to sentences, and classify unknown sentences by assigning them one of the learned stance labels. However, building supervised machine learning structures on top of manual stance annotation imposes limitations on the model's ability to fully generalize across different contexts. 

The advent of Large Language Models (LLMs) opens up the possibility of classifying the stance of a huge number of sentences with an absolute minimum use of human labels. This allows for unsupervised stance detection, thus curtailing the need for manual stance annotation. With the lighting speed of LLM research, there are a plethora of LLMs released, and a variety of prompting schemes for formatting the input into the LLMs. In this paper, we study the use of LLMs for the stance detection task, essentially investigating the following questions. How suitable are LLMs for the stance detection task? Which model produces the best performance? How should one format the inputs to the model? To answer these questions, we perform the stance classification task across 10 LLMs and 7 prompting schemes. To ensure generalizability of our results, we perform our experiments across 6 benchmark, social media datasets that have been manually annotated for their stance.

\subsection{Objectives and Contributions}
This article studies the stance classification task with social media datasets, namely Twitter datasets. The stance classification task for social media datasets is particularly important in today's digital age. Understanding the stance of social media conversations is key towards societally critical tasks such as distilling the aggregate public opinion towards important or contentious topics\cite{alkhalifa2022capturing}.

Within this work, we test Large Language Models as a stance classification technique. LLMs provide an easy and intuitive way of extracting information from a sentence. Instead of having to code or train a particular model for a particular task, an analyst can simply type a prompt in natural language for the LLM to perform that task. This ease of analysis opens up text analysis to a variety of analyst types, from computational to social to policy analysts. A zero-shot inference scheme also aids in extracting stance from the volume of social media posts, for no prior training of a language model is required for many language tasks with the current state-of-the-art LLMs. However, for the best performance of LLMs, including accurately extracting a stance from a sentence, users and analysts must be able to formulate the appropriate prompts. Therefore, in this paper, we investigate different prompting schemes for LLMs for stance classification. Finally, since LLMs can be fine-tuned for a task to further potentially boost performance of the LLM for that task, we also investigate how fine-tuning affects LLM performance for stance classification.

The key objective of this article is to investigate the ability of LLMs for the task of stance detection, with a focus on different prompting schemes. We performed experiments on several social media datasets to provide insight into how the performance of LLMs in the stance detection task varies with the context of the data.

Our main contributions can be summarized as follows.
\begin{itemize}
    \item We evaluated the stance detection task using Large Language Models (LLMs). Our experiments span 6 Twitter datasets, 10 models, and 7 prompting schemes. This large-scale evaluation reveals that LLMs are often competitive with in-domain supervised models but do not consistently outperform these supervised baselines. LLMs tend to outperform supervised models only with specific prompting schemes, indicating the importance of crafting an input prompt.
    \item To analyze whether specialized models perform better, we experimented on fine-tuned LLMs. Our results show that fine-tuned models do not necessarily perform better than zero-shot models.
    \item We analyzed the responses of LLMs when instructed to predict a stance from a sentence. We find that for those data points where the LLM produced a valid stance classification for the statement to be classified, about half of those entries (50.8\%) were correctly predicted.
\end{itemize}

In the rest of this article, we review the relevant studies in \autoref{sec:related} in terms of the stance detection task, stance detection in social networks, and the use of Large Language Models for stance detection. We next detail the datasets prepared for the stance detection task in \autoref{sec:datasets}. The stance detection methodology is detailed in \autoref{sec:method}, where we describe the LLMs tested, the crafting of the prompting schemes, and the fine-tuning methodology. We present the results of the experimental evaluations quantifying the ability of LLMs to detect stance in \autoref{sec:results}. \autoref{sec:discussion} discusses the results and outlines directions for future work before \autoref{sec:conclusion} concludes the article.

\section{Related Work}
\label{sec:related}
\subsection{Stance Classification Task}
Stance classification is a well-studied natural language processing task. At its core, stance classification is the task of classifying an entity's (e.g., person, text statement, etc.) opinion, either implied or expressed, about a target \cite{ng2022my}. As a result, this task has an inherent context-dependency in its definition that related tasks, unlike sentiment which do not depend on the context: stance is not defined without a target, and so classifying stance requires this context. Due to the importance of determining stance for many disciplines, this task has several benchmark datasets and methods. In previous work involving stance classification, much of the focus was on the construction of supervised machine learning models for the task. A commonly used machine learning classifier is the Support Vector Machine \cite{lai2018stance,elfardy2016cu}, which has performed well in the SemEval-2016 benchmark stance detection competition \cite{mohammad2016semeval}. Supervised models that use neural network architectures are also popular. These include the use of convolutional neural networks \cite{wei2016pkudblab}, recurrent neural networks \cite{zarrella2016mitre}, and the enhancement of these network architectures with textual entailment \cite{zhao2020pretrained} and data augmentation \cite{kawintiranon2021knowledge}. Other recent approaches focus on multi-task learning objectives and transfer learning from transformer-based neural networks \cite{alturayeif2023systematic,yang-etal-2019-blcu,zhao2020pretrained}. The resultant stance models typically boast strong in-domain performance, but they often struggle to generalize to new data or other unseen stance targets and thus are of little use to real world practitioners due to this generalizability shortcoming \cite{ng2022my,alturayeif2023systematic}.

Although most of the work in stance classification has focused on supervised machine learning models built upon manual human annotations, there are also some unsupervised techniques. Unsupervised stance labeling methods make use of the idea of language homogeneity for label classification \cite{zhang2023doubleh}. One popular technique is classification using graph networks. Graph neural networks have been used to formulate homogeneous and heterogeneous information for Twitter users, thereafter inferring the stance of the user based on past tweets and the tweets of its interaction graph neighbors \cite{zhang2023doubleh}. Another technique is the use of label propagation on user interaction networks that represent relationships between users \cite{10.1145/2492517.2492557}. The interaction network can also be divided into partitions, and the user stance can be interpreted as an aggregated stance of partitions \cite{pick2022stem}. \citet{darwish2020unsupervised} projected a large set of Twitter user data onto a low-dimensional space before clustering the interaction network into multiple partitions. These unsupervised methods do not rely on having a set of stance labels to train a model but often have very specific scenarios for their use, such as the explicit behavioral links between social media users to create a graph network. It also relies on strong assumptions about the language or behavior of users to infer stances, such as the association of a particular stance towards a word or hashtag.

A last family of techniques in the stance classification task is the use of zero-shot methods for stance detection. Zero-shot classification is the idea that the model can classify items that it has not encountered previously without being trained on a particular use case.  In particular, \cite{allaway2023zero} discuss a variety of techniques for zero-shot stance detection and present an adaptation of the SemEval2016 dataset \cite{zarrella2016mitre}, along with their own VAST dataset, specifically designed for the purpose of using a zero-shot technique for stance classification. They highlight the multitude of ways that stance characterization can be achieved in a zero-shot manner. Essentially, there are three primary paradigms for zero-shot stance detection: topic, language, and genre \cite{allaway2023zero}. For each paradigm, a model is trained via an out-of-domain setup: a model is trained on all data except for one element of the paradigm, which is reserved for the zero-shot test. For instance, in a zero-shot topic stance detection, the model is trained on data from all topics except one, which is then used for evaluation. All models perform less optimally in the zero-shot setting compared to a fully supervised setting \cite{allaway2023zero}.

\subsection{The use of Large Language Models for Stance Classification}
Large Language Models (LLMs) have the capability to process and evaluate sentences of natural languages, and have been used for tasks ranging from reading comprehension to mathematical problem solving \cite{zheng2023judging}. These models are trained on an enormous amount of data and have an uncanny ability to evaluate sentences and respond to input prompts. Some research has emerged on using these models for the stance classification task. \autoref{tab:stance_llm} lists a table of current works involving stance detection using LLMs.

Although a suite of LLMs has been developed, much of the current work has focused solely on the GPT family (i.e., ChatGPT, GPT3) \cite{chatGPT}, producing mixed results. \cite{zhang2022would} found that with just an instruction-based prompt (although they hint at the use of reasoning in a prompt in their paper), ChatGPT could produce better results on the SemEval2016 benchmark data set than supervised models. On the other hand, \cite{aiyappa2023can} examined the stance detection task using the ChatGPT model and observed that while there is a performance boost, there could be potential data contamination through its massive training dataset, therefore making the evaluation unreliable. Furthermore, \cite{mets2023automated} investigated the usability of ChatGPT as a zero-shot stance detection classifier, using only a task-based prompt for stance detection on a custom dataset for the topic of immigration in news articles for various languages. In their experiments, they found that ChatGPT performed close to the best supervised model, but was ultimately inferior for the stance classification problem. Most recently, a series of papers have proposed different Chain-of-Thought style prompts for stance classification, generally showing superior performance when the reasoning abilities of LLMs are deliberately invoked by these prompting schemes \cite{zhang2023logically,zhang2023investigating}. Finally, and most recently, \cite{lan2023stance} used a multistage Tree-of-Thought-like prompt that also employs embodiements to deliver state-of-the-art performance on a couple of benchmark stance datasets for both GPT and open-source models.

Also, given the similarities between stance and sentiment classification, it is worth mentioning that \cite{kheiri2023sentimentgpt} investigated several OpenAI model offerings on the sentiment classification task and found that GPT models, especially if fine-tuned, can outperform other models. Overall, it is unclear if LLMs can perform the task of stance classification, especially with some prompt engineering and without the use of the additional step of model fine-tuning.

LLMs take in an input, which is called a prompt, which provides instructions to the model to provide a certain output. With the prompt being a free text format, a new discipline of prompt engineering has come into being. Prompt engineering is a discipline that tries to find the right ways to provide input to the LLMs -- or ``prompt" them -- to get the best outputs \cite{schmidtcataloging, white2023prompt, Ramlochan2023prompt}. While this is a fast changing discipline, a couple of patterns have emerged. The first is zero-shot prompting, where the LLM is provided only the task description and asked for the desired output. Next we have few-shot prompting, where the LLM is given a few examples of the desired output as part of the prompt \cite{white2023prompt, brown2020language, wei2023larger}. This is different from fine-tuning the LLM or low-shot learning, as no training (i.e. adjusting of model weights) is performed when the examples are given; the examples are only given as part of the context to aid understanding of the task \cite{brown2020language}. While this prompting technique consistently produces improved outputs, there are still possible instabilities in the technique that can be caused by things such as the order of the provided examples \cite{zhao2021calibrate, lu2021fantastically}. Another prompting technique that has consistently improved the output of LLMs is Chain-of-Thought Reasoning \cite{wei2022chain, chen2023you}. In this prompting scheme, the LLM is asked to explain its reasoning and to work through answering a prompt step by step. Answering prompts in such a process usually improves outcomes and helps to prevent undesirable behaviors like hallucination, where the model produces a plausible-looking answer that is incorrect \cite{wei2022chain, chen2023you}. This technique has been used in an iterative chat format to improve implicit sentiment classification in previous research \cite{fei2023reasoning}. Although there are certain prompting techniques that can elicit better outcomes from LLMs, the best means of interacting with the LLMs, especially in the domain of stance detection, is still an open research question.

\begin{table}
    \centering
    \begin{tabular}{|p{3cm}|p{4cm}|p{3cm}|p{4cm}|}
        \hline
        \textbf{Work} & \textbf{Dataset} & \textbf{Base LLM} & \textbf{Prompting Scheme} \\ \hline
        \citet{aiyappa2023can} & SemEval2016\cite{mohammad2016semeval} & ChatGPT & Task-based prompt with target\\ \hline
        \citet{kheiri2023sentimentgpt} & SemEval2017 & GPT-3, GPT-3, GPT-3.5 & Embodied task-based prompt with definitions and target prompt \\ \hline
        \citet{lan2023stance} & SemEval2016\cite{mohammad2016semeval}, VAST\cite{allaway2020zero}, wtwt\cite{wtwt} & GPT-3.5 & Embodied Tree-of-Thought prompt \\ \hline
        \citet{liyanage2023gpt} & Climate change & GPT-4 & Zero-shot and few-shot Chain-of-Thought prompts \\ \hline 
        \citet{li2023stance} & SemEval2016\cite{mohammad2016semeval}, P-Stance, COVID19 & ChatGPT & Discourse expanding prompt and few-shot prompt \\ \hline 
        \citet{mets2023automated} & Immigration in news articles & ChatGPT & Task-based prompt \\ \hline
        \citet{vsuppa2024bryndza} & CASE2024 Shared Task Climate Activism Stance and Hate Event Detection \cite{thapa2024stance} & GPT-4, LLaMA & Few-shot prompt \\ \hline 
        \citet{zhang2022would} & SemEval2016\cite{mohammad2016semeval} & ChatGPT & instruction-based prompt \\ \hline 
        \citet{zhang2023logically} & SemEval2016\cite{mohammad2016semeval}, VAST\cite{allaway2020zero} & GPT-3 & Logically-Consistent Chain-of-Thought prompt \\ \hline
        \citet{zhang2023investigating} & SemEval2016\cite{mohammad2016semeval}, VAST\cite{allaway2020zero}, P-Stance & ChatGPT & One-shot Chain-of-Thought prompt \\
        \hline
    \end{tabular}
    \caption{Current Work in Stance Detection using Large Language Models}
    \label{tab:stance_llm}
\end{table}

\subsection{Stance Classification for Social Media}
Stance classification is often used for social media analysis studies. Social media platforms are a major component of a person's social interaction and are tools for dissemination of information to express opinions \cite{aldayel2021stance}. Stance detection on social media provides a conduit for measuring public opinion, especially on controversial political and social issues. Such issues can be topics like abortion and climate change \cite{mohammad2016semeval}, or support and opposition to political debates on online forums \cite{murakami2010support}. Other analyses involving stances on social media include observing the change in stance toward topics such as mandatory vaccination \cite{ng2022pro}, or the role of automated and malicious actors in polarizing stances on a certain topic \cite{aldayel2022characterizing}.

Stance classification for social media texts is a complicated problem. There are two key challenges in classifying stance for social media texts: a general lack of human-annotated data and the huge volume of text present in any given social media space. Stance classification via LLMs can provide respite to analysts for the labeling process and could potentially be a zero-shot classification process and not require further model training or additional annotations.

\section{Data Sets}
\label{sec:datasets}
In this work, we used a total of six publicly available benchmark datasets: covid-lies \cite{covidlies}, election2016 \cite{sobhani2017dataset}, phemerumors \cite{phemerumours}, SemEval2016 \cite{mohammad2016semeval}, srq \cite{villa2020stance}, and wtwt \cite{wtwt}. These six data sets have similar properties: they are constructed from sentences that are Twitter posts, are written in the English language, and were manually annotated by human labelers. The targets of these data sets range from misconceptions to elections to a root tweet in a reply chain, which means the definition of \textit{stance} varies between them. For example, in covid-lies and phemerumors the stance is about whether the statement supports or denies a rumor, while in SemEval2016 and election2016 the stance is about an opinion of the politically-divisive target. \autoref{tab:datasets} lists the datasets that are used, the targets that are present in the data sets, and the highest unweighted F1 score accuracy reported as found in the original articles. We urge the reader to refer to the original dataset papers for a deeper understanding of the dataset contents.

\begin{table*}[!ht]
\centering
\begin{tabular}{|l|l|l|l|}
\hline
\textbf{Data Set} & \textbf{Targets}   & \textbf{Unweighted F1-score} & \textbf{Number of Examples}  \\ \hline
\textbf{covid-lies} & misconceptions towards COVID-19 pandemic & 0.50  & 3,196   \\ \hline
\textbf{election2016} & 2016 US Presidential elections & 0.55  & 2,378   \\ \hline
\textbf{phemerumors}  & tragedies (unrest, disasters, hostage, plane crash)  & 0.33  & 2,859    \\ \hline
\textbf{SemEval2016} & atheism, climate change, feminism, Hillary Clinton, abortion & 0.69  & 2,814    \\ \hline
\textbf{srq} & student marches, school shooting, Iranian nuclear deal & 0.48 & 4,215 \\ \hline
\textbf{wtwt} & company mergers and acquisitions   & 0.62  & 32,409   \\ \hline
\end{tabular}
\caption{Summary of data sets used with our descriptions of the events and best-reported unweighted, macro F1-score from the original data set.}
\label{tab:datasets}
\end{table*}

\section{Methodology}
\label{sec:method}
In this section, we detail the methodology we used to classify the stances by LLMs. We first elaborate on the architecture and construction of LLMs used, before detailing the series of prompting techniques we tested in this work. We then described how we fine-tuned the LLMs for stance classification. All of the experiments were run on a computer with Ubuntu 22.04 Linux with x64 CPU with 40 cores, 376 GB of RAM and four NVIDIA A6000 GPUs.

\subsection{Large Language Models Used}
In this work, we tested 10 Large Language Models (LLMs). These models varied in architecture and sizes. We also tested the same model architecture with different sizes (e.g., Llama-2 13B Chat vs Llama-2 7B Chat) or pretrainings (e.g., Llama-2 7B Chat vs Llama-2 7B). We used the HuggingFace version\footnote{\url{https://huggingface.co/}} of the LLMs. We selected open-source LLMs that can be reliably run on a high-end computer, mimicking how a social media analyst would run stance detection for social media datasets.  The following are the LLMs that we tested throughout this study: 
\begin{enumerate}
    \item \textbf{Falcon 7B Instruct:} \cite{shazeer2019fast} Falcon is a series of decoder-only models. Relative to other decoder-only models the Falcon series uses different architectures, including Flash Attention. This variant has 7 billion parameters, fine-tuned on instruction datasets.
    \item \textbf{Falcon 40B Instruct:} \cite{shazeer2019fast} This variant of the Falcon architecture has 40 billion parameters and is fine-tuned on instruction datasets. 
    \item \textbf{T5 3B Flan-Alpaca:} \cite{chung2022scaling} Flan-Alpaca combines two elements: the alpaca direction of fine-tuning to approximate LLMs for cheaper and easier training and the encoder-decoder T5 model. This variant has 3 billion parameters.
    \item \textbf{T5 11B Flan-Alpaca:} \cite{chung2022scaling} This Flan-Alpaca tuned T5 variant has 11 billion parameters.
    \item \textbf{Mistral v0.1 7B Instruct:} \cite{jiang2023mistral} Mistral is a decoder-only model, similar to Falcon or Llama-2, but with different architectural features like Grouped-Query Attention and Sliding Window Attention. The Mistral model is 7 billion parameters and has been instruction tuned.
    \item \textbf{UL2 20B Flan:} \cite{tay2022unifying} UL2 20B Flan is an encoder-decoder model of 20 billion parameters based on the T5 architecture and is instruction tuned.
    \item \textbf{Llama-2 13B Chat:} \cite{touvron2023llama} Llama-2 is a decoder-only model that incorporates elements such as a different token encoding model and Flash Attention 2. This version of the Llama series of models has 13 billion parameters and has been tuned on chat datasets.
    \item \textbf{Llama-2 7B Chat:} \cite{touvron2023llama} This version of the Llama-2 architecture has 7 billion parameters and is optimized for dialogue use cases.
    \item \textbf{Llama-2 7B:} \cite{touvron2023llama} This version of the Llama-2 architecture has 7 billion parameters, but has not been further tuned beyond pre-training of the model.
    \item \textbf{Phi-2:}\cite{gunasekar2023textbooks} Designed by Microsoft, Phi-2 is a decoder-only transformer-based LLM with 2.7 billion parameters. It is trained on Python codes, synthetic texts and website content. 
\end{enumerate}

For each of the models, we use greedy decoding to in order to get consistency across models and replicability in results. We also experimented with other forms of decoding, such as beam search and top-p, with Mistral and Llama models on the election and SemEval2016 datasets, but did not find major performance differences between what these models produced with just greedy decoding. 

\subsection{Prompting Schemes for Stance Classification}
\begin{figure*}[!ht]
    \centering
    \includegraphics[width=\textwidth]{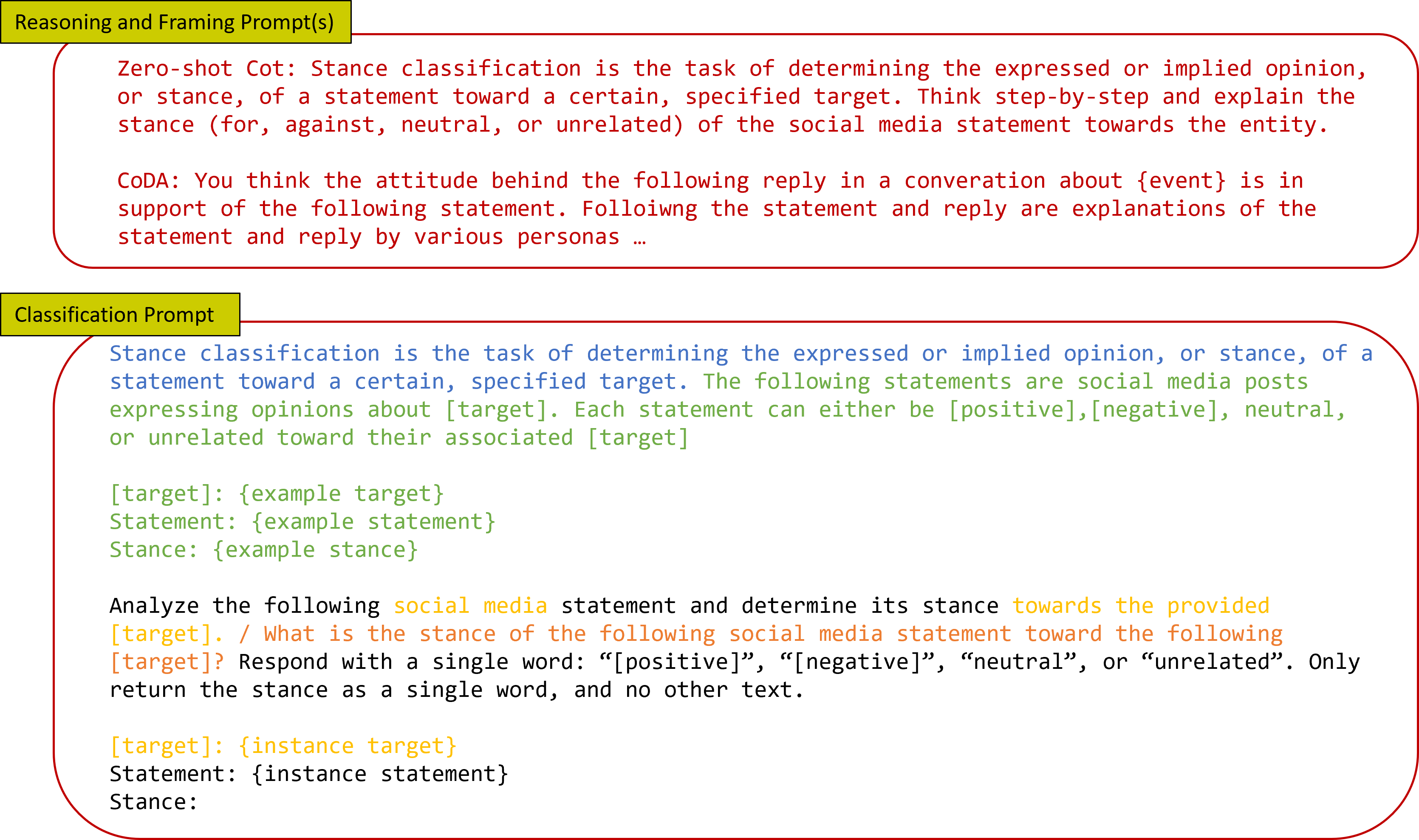}
    \caption{Overarching Prompting Scheme for Stance Classification 
    \textmd{Text highlights indicate the information available for each of the prompting schemes. Black text indicates the basic task-only prompt, whose elements are common to all of the other prompts. Blue indicates the addition of the task definition. Yellow indicates the addition of context, whereas the orange color is the context question variation of the context prompt. Green indicates the prompt additions from few-shot prompting. Finally, red indicates the additional reasoning or framing prompts introduced in prompting methods like CoT and CoDA that attempt to elicit preliminary reasoning about the statement before trying to classify the stance of the statement. Those words bracketed in square brackets are dataset dependent, while those bracketed in curly braces are instance-dependent.}}
    \label{fig:prompt_template}
\end{figure*}

In order to investigate the use of LLMs and prompting for the task of stance classification, we used seven different prompting schemes. The prompting schemes are generally hierarchical so that each prompting scheme incorporates more information from the previous scheme. \autoref{fig:prompt_template} displays the general, overarching prompting scheme and the elements available to the LLMs within each individual prompting scheme. Actual examples of each prompt scheme are provided in the \autoref{sec:prompt_examples}. 

The following list details each of the prompting schemes we used to classify stance for each of the data sets. In a stance detection task, the input sentence comes with a target to which it is expressing opinion towards and stance options. \autoref{tab:prompt_details} describes the different classification outcomes of the datasets used in this study. The core text of the prompts are tweaked towards the properties of each dataset.

\begin{enumerate}
    \item \textbf{Task-only:} In the task-only prompt, we adopt a zero-shot learning prompting method, providing only the task with no additional detail (e.g. \texttt{``Analyze the following statement and determine its stance.``}). For this prompt, we provide different classification outcomes depending on the data set \autoref{tab:prompt_details}), but they generally follow a \texttt{`agree'}, \texttt{`disagree'}, or \texttt{`neutral'}// \texttt{`unrelated'} format, with a final output instruction to only output the label. Thus, the remainder of the prompt is: \texttt{Respond with a single word: ``for", ``against", ``neutral", or ``unrelated". Only return the stance as a single word, and no other text.}
    
    \item \textbf{Task Definition:} In the task definition prompt, we adopt the zero-shot learning prompting method, but now provide the task and the definition of `stance classification' (e.g. \texttt{ ``Stance classification is the task of determining the expressed or implied opinion, or stance, of a statement toward a certain, specified target. Analyze the following statement and determine its stance.``}) similar to what was done in \cite{mets2023automated}. The outcome labels and output instructions remain the same as in the task-only prompt.
    
    \item \textbf{Context Analyze: } In this prompting scheme, we add contextual information about what the statement is and the target of the stance classification to the task of classifying the stance of the statement (e.g.  \texttt{``... Analyze the following social media statement and determine its stance towards the provided [target]. ...``}). We keep the definition of stance classification as previously given in the task definition prompt. This prompting scheme is what the previous works that investigated using ChatGPT for stance labeling used \cite{zhang2022would, aiyappa2023can, mets2023automated}. The outcome labels and output instructions remain the same as in the task-only prompt.
    
    \item \textbf{Context Question: } This prompting scheme is the same as the previous one, but with one variation in the task part of the prompt. Instead of phrasing the task as a statement with `analyze', we instead phrase it as a question. In our review of the literature, we noticed several ways of prompting the LLM for stance some of which included prompting by questioning \cite{zhang2022would}. In this prompting scheme we change the task sentence to  \texttt{``... What is the stance of the following social media statement towards the following [target]? ...``}. Other than this change, all other aspects of the prompt remain the exact same as those of the context analyze prompt.
    
    
    \item \textbf{Zero-shot CoT:} For this prompting scheme, we adopt a Chain of Thought prompt to elicit reasoning from the LLM as was done in \cite{liyanage2023gpt, zhang2023investigating}. In this case, there are two prompts, one to have the LLM reason about the stance and for LLM to evaluate its reasoning and then make a final stance determination. The reasoning prompt is given by \texttt{``Stance classification is the task of determining the expressed or implied opinion, or stance, of a statement toward a certain, specified target. Think step by step and explain the stance (for, against, neutral, or unrelated) of the social media statement towards the [target]."} The results of this prompt are then fed into the final stance determination prompt as: \texttt{``Therefore, based on your explanation, [reason prompt results], what is the final stance? ...} with the same outcome labels and output instructions as in the task-only prompt.

    \item \textbf{CoDA:} Finally, we also try the role-infused collaborative agents prompt (CoDA) presented in \cite{lan2023stance}. For this prompting scheme, there are six total prompts: The first three prompts are for the analyses by the different embodiments (i.e., linguistics expert, subject matter expert, and heavy social media user). Then there are two prompts for evaluating the experts and forming arguments; one prompt for forming the positive or for argument, and one prompt for forming the against argument. Then, finally, there is one last evaluation prompt that evaluates the arguments and outputs the final stance label. As with the previous prompts, the output labels and output instructions remain the same. However, unlike the previous prompts, this prompt does not use the definition of stance as part of the prompt. In general, this prompting scheme combines elements of Tree-of-Thought prompting \cite{yao2023tree}, by positing arguments or ideas and then evaluating those arguments or ideas, with embodied prompting where the LLM is given a role to play or embody.
\end{enumerate}

\begin{table*}[!ht]
\begin{tabular}{|l|p{5cm}|l|l|}
\hline
\textbf{Data Set} & \textbf{Stance Options} & \textbf{Target} \\ \hline
\textbf{covid-lies} & supports, denies, neutral, unrelated & belief about COVID \\ \hline
\textbf{election2016} & for, against, neutral, unrelated & politician \\ \hline
\textbf{phemerumors} & supports, denies, neutral, unrelated & rumor \\ \hline
\textbf{SemEval2016}  & for, against, neutral, unrelated  &  entity \\ \hline
\textbf{srq} & supports, denies, neutral, unrelated & social media statement about an event \\ \hline
\textbf{wtwt} & supports, denies, neutral, unrelated & corporate merger happening \\ \hline
\end{tabular}
\caption{Summary of the prompt differences between each of the benchmark data sets. For each data set, we used slightly different target and stance label options in order to accommodate the different purposes of stance classification between the data sets.}
\label{tab:prompt_details}
\end{table*}

\subsection{Fine-Tuning an LLM for Stance Classification}
A popular way of adapting language models to tasks is through the fine-tuning process, which molds these models into specialized ones to analyze niche topics and achieve better performance \cite{ding2023parameter,hu2021lora}. To investigate whether a specialized model can perform better for the stance classification task, we performed fine-tuning on some exemplar LLMs. Due to computational resource limitations, we selected representative LLMs of each type and of different sizes to fine-tune. For encoder-decoder models, we used: T5 11B Flan-Alpaca and UL2 20B Flan. For decoder-only models, we used: Mistral v0.1 7B Instruct, Llama-2 7B and Llama-2-7B-Chat.

To fine-tune the models, we opted to use the parameter-efficient method of LoRA \cite{hu2021lora}. We used a parameter-efficient technique (versus full fine-tuning) as we only have a handful of benchmark datasets to train on for this task, and LLMs have a massive number of parameters. Furthermore, recent research has found LoRA to be more stable in fine-tuning and capable of providing good fine-tuning results in scenarios with limited training data \cite{zhang2023scaling}. Specifically, we follow the guidance set out in \cite{hu2021lora} and use a LoRA adapter on all the LLM attention layers with default parameters of $r=16$, $\alpha=1$, and a LoRA dropout of $0.1$. We trained these models with a learning rate of $1e-5$ with a linear learning rate decay and 100 warmup steps, for a maximum of three epochs, using the AdamW optimizer \cite{loshchilov2017decoupled} and mixed precision training \cite{micikevicius2017mixed}. For the training task, we evaluated whether the model outputs the correct label, in a casual language modeling scenario. In other words, the model could output any tokens, but would only be correct if output the tokens for the correct stance classification and only those tokens.

We fine-tuned the models using a leave-one-out approach across the datasets. This tests how well the model performs on out-of-domain datasets, which is critical for the stance classification task. This setup also mimics the zero-shot models, which are not specialized for the stance classification task nor its datasets, and performs classification based on pre-trained knowledge only. This idea is adopted from previous work that demonstrated that aggregating multiple stance datasets improves out-of-context stance detection, since the model has more data to work with creating a generalized stance classifier \cite{ng2022my}. Therefore, we iteratively hold out one dataset as the test dataset, and we fine-tuned the LLM on all the other datasets except the test dataset. We allowed the models to stop training when they had minimzed their loss on the held-out dataset (early stopping with 5 steps as patience). Then we perform stance detection on the test dataset and we report the F1 accuracy score on the held-out dataset. In this way, we can get a measure of how well fine-tuning works and make a comparison between fine-tuned models to simply prompting a pre-trained LLM. We encapsulated each datum in the Context Analyze prompt and fed the encapsulated datum into the model during the fine-tuning stage. The same prompt format was used to label the held-out test dataset after fine-tuning.

\subsection{Evaluation Metrics}
Each of the datasets have different sets of labels of stances. To make the evaluation comparable across the datasets, we followed the same data set handling procedures described in \cite{ng2022my} and standardized the stance labels for evaluation. This standardization step maps the set of stance labels for each dataset to a common set of \{agree, neutral, disagree\} labels. Note that the stance labels are not standardized in the prompts during the experimental stage, because each dataset is constructed to elicit the opinions towards different targets and we aim to preserve the definition of stance towards each of the varied targets.

For each response obtained from the LLMs, we further post-process the response. This is required because, despite instructing the models to only return a single stance label, the models sometimes return extra text, including explanations. We construct lists of common labels of \{agree, disagree, neutral\} that the models return for each stance and match them within the returned response. If the returned response contains words from only one list, that data point is classified as that stance category. If more than one category word is detected from multiple categories or the model fails to output a stance label, the stance of the data point is labeled as ``neutral". The data points that have a valid stance label as previously described (i.e., those with one and only one stance label present in the output) are termed as ``good results", while those that have no valid stance are termed as ``bad results". For example, `agree' and `the stance is agree' are valid labels and thus are good results, whereas `unconfirmed' and `agree, neutral, disagree' are not. We report the aggregated F1 scores of all data points as they are resolved, and also present the breakdown between good and bad results.

For evaluation of the stance detection task against the manually annotated gold labels, we report the unweighted, macro-F1 accuracy metric, in line with previous work \cite{mohammad2016semeval}. The macro-F1 score adjusts for the proportion of each class label type (i.e., stance label), for there is an imbalance of class labels in some of our datasets. We use the accuracy scores of the original dataset models as a baseline comparison.

\section{Results}
\label{sec:results}
In this section, we present the results of different prompting schemes and LLM combinations, the results of the fine-tuning models and observations on the quality of the LLM responses. We present a graphical visualization of all these results in \autoref{sec:graphical_results}.

\subsection{Results of LLM \& Prompting Schemes Combinations}
\autoref{tab:results} presents the results of the combinations of LLMs, prompting schemes and datasets in terms of the unweighted F1-scores. From our results, we observe that while some combinations of LLMs and prompting schemes outperform the baselines, not all combinations consistently outperform the baselines. This speaks to the inconsistency and random nature of LLM inference, which can reduce the consistency and performance of using LLM as a stance classification technique \cite{xu2024unilog}. 

In general, Few-Shot Prompting (FSP) and Zero-Shot CoT (CoT) perform better, as do the encoder-decoder family of models.
Unfortunately, there is no one LLM that consistently outperforms across all the datasets, nor is there an LLM that consistently outperforms all other models regardless of prompting schemes in a single dataset. Similarly, there is no one prompting scheme that outperforms all other prompting schemes in this stance detection task.
In addition, a larger model size does not necessarily mean that the model performs better. For example, Falcon 40B Instruct vs. Falcon 7B Instruct with the FSP scheme shows a performance of 0.25 against 0.35.

\begin{longtable}{|l|l|l|p{1.5cm}|p{1.5cm}|l|p{1.5cm}|l|}
    \hline
        \textbf{LLM/Prompting Scheme} & \textbf{Task-Only} & \textbf{Task Definition} & \textbf{Context Analyze} & \textbf{Context Question} & \textbf{Few Shot}  & \textbf{Zero-shot CoT} & \textbf{CODA} \\ \hline
        \textbf{covid-lies} & ~ & ~ & ~ & ~ & ~ & ~ & ~ \\ \hline
        Baseline & 0.50 & ~ & ~ & ~ & ~ & ~ & ~ \\ \hline
        Falcon 40B Instruct & 0.19 & 0.19 & 0.23 & 0.29 & 0.25 & 0.36 & ~ \\ \hline
        Falcon 7B Instruct & 0.32 & 0.31 & 0.14 & 0.14 & 0.35 & 0.23 & 0.29 \\ \hline
        T5 3B Flan-Alpaca & 0.17 & 0.17 & 0.33 & 0.31 & 0.40 & 0.34 & 0.24 \\ \hline
        T5 11B Flan-Alpaca & 0.22 & 0.21 & 0.36 & 0.34 & 0.49 & 0.41 & 0.4 \\ \hline
        Mistral v0.1 7B Instruct & 0.23 & 0.28 & 0.42 & 0.38 & 0.48 & 0.37 & 0.33 \\ \hline
        UL2 20B Flan & 0.13 & 0.16 & 0.5 & 0.51 & \textbf{0.52} & \textbf{0.52} & 0.28 \\ \hline
        Llama-2 13B Chat & 0.27 & 0.23 & 0.27 & 0.31 & 0.36 & 0.24 & 0.19 \\ \hline
        Llama-2 7B Chat & 0.27 & 0.29 & 0.19 & 0.2 & 0.28 & 0.12 & 0.18 \\ \hline
        Llama-2 7B & 0.29 & 0.26 & 0.22 & 0.17 & 0.34 & 0.23 & ~ \\ \hline
        Phi-2 & 0.32 & 0.3 & 0.29 & 0.15 & 0.27 & 0.32 & 0.31 \\ \hline
        \textbf{LLM/Prompting Scheme} & \textbf{Task-Only} & \textbf{Task Definition} & \textbf{Context Analyze} & \textbf{Context Question} & \textbf{Few Shot}  & \textbf{Zero-shot CoT} & \textbf{CODA} \\ \hline
        \textbf{election2016} & ~ & ~ & ~ & ~ & ~ & ~ & ~ \\ \hline
        Baseline & 0.55 & ~ & ~ & ~ & ~ & ~ & ~ \\ \hline
        Falcon 40B Instruct & 0.38 & 0.41 & 0.48 & 0.43 & 0.51 & 0.47 & ~ \\ \hline
        Falcon 7B Instruct & 0.18 & 0.26 & 0.27 & 0.29 & 0.31 & 0.32 & 0.27 \\ \hline
        T5 3B Flan-Alpaca & 0.34 & 0.32 & 0.32 & 0.32 & 0.42 & 0.41 & 0.4 \\ \hline
        T5 11B Flan-Alpaca & 0.43 & 0.38 & 0.48 & 0.46 & \textbf{0.60} & 0.49 & 0.48 \\ \hline
        Mistral v0.1 7B Instruct & 0.42 & 0.43 & 0.49 & 0.48 & 0.41 & 0.5 & 0.31 \\ \hline
        UL2 20B Flan & 0.4 & 0.41 & 0.46 & 0.49 & 0.5 & 0.37 & 0.5 \\ \hline
        Llama-2 13B Chat & 0.38 & 0.38 & 0.41 & 0.32 & 0.52 & 0.31 & 0.33 \\ \hline
        Llama-2 7B Chat & 0.38 & 0.38 & 0.39 & 0.36 & 0.49 & 0.3 & 0.36 \\ \hline
        Llama-2 7B & 0.27 & 0.31 & 0.24 & 0.32 & 0.24 & 0.33 & ~ \\ \hline
        Phi-2 & 0.28 & 0.24 & 0.28 & 0.3 & 0.33 & 0.32 & 0.24 \\ \hline
        \textbf{LLM/Prompting Scheme} & \textbf{Task-Only} & \textbf{Task Definition} & \textbf{Context Analyze} & \textbf{Context Question} & \textbf{Few Shot}  & \textbf{Zero-shot CoT} & \textbf{CODA} \\ \hline
        \textbf{phemerumors} & ~ & ~ & ~ & ~ & ~ & ~ & ~ \\ \hline
        Baseline & 0.33 & ~ & ~ & ~ & ~ & ~ & ~ \\ \hline
        Falcon 40B Instruct & 0.16 & 0.15 & 0.16 & 0.22 & 0.14 & 0.20 & ~ \\ \hline
        Falcon 7B Instruct & 0.23 & 0.26 & 0.13 & 0.13 & 0.15 & 0.17 & 0.23 \\ \hline
        T5 3B Flan-Alpaca & 0.13 & 0.13 & 0.22 & 0.21 & 0.23 & 0.22 & 0.14 \\ \hline
        T5 11B Flan-Alpaca & 0.16 & 0.15 & 0.22 & 0.22 & 0.26 & 0.24 & 0.26 \\ \hline
        Mistral v0.1 7B Instruct & 0.21 & 0.23 & 0.29 & 0.2 & 0.26 & 0.23 & 0.19 \\ \hline
        UL2 20B Flan & 0.15 & 0.15 & 0.3 & 0.30 & 0.31 & \textbf{0.29} & 0.19 \\ \hline
        Llama-2 13B Chat & 0.21 & 0.20 & 0.23 & 0.25 & 0.15 & 0.19 & 0.19 \\ \hline
        Llama-2 7B Chat & 0.21 & 0.24 & 0.20 & 0.21 & 0.18 & 0.14 & 0.15 \\ \hline
        Llama-2 7B & 0.23 & 0.20 & 0.22 & 0.18 & 0.24 & 0.19 & ~ \\ \hline
        Phi-2 & 0.24 & 0.27 & 0.10 & 0.07 & 0.25 & 0.19 & 0.23 \\ \hline
        \textbf{LLM/Prompting Scheme} & \textbf{Task-Only} & \textbf{Task Definition} & \textbf{Context Analyze} & \textbf{Context Question} & \textbf{Few Shot}  & \textbf{Zero-shot CoT} & \textbf{CODA} \\ \hline
        \textbf{SemEval2016} & ~ & ~ & ~ & ~ & ~ & ~ & ~ \\ \hline
        Baseline & 0.69 & ~ & ~ & ~ & ~ & ~ & ~ \\ \hline
        Falcon 40B Instruct & 0.40 & 0.40 & 0.60 & 0.58 & 0.52 & 0.44 & ~ \\ \hline
        Falcon 7B Instruct & 0.40 & 0.4 & 0.24 & 0.23 & 0.36 & 0.28 & 0.33 \\ \hline
        T5 3B Flan-Alpaca & 0.40 & 0.40 & 0.54 & 0.54 & 0.6 & 0.6 & 0.56 \\ \hline
        T5 11B Flan-Alpaca & 0.4 & 0.4 & \textbf{0.64} & 0.63 & 0.63 & 0.58 & 0.61 \\ \hline
        Mistral v0.1 7B Instruct & 0.42 & 0.39 & 0.51 & 0.53 & 0.47 & 0.5 & 0.31 \\ \hline
        UL2 20B Flan & 0.4 & 0.41 & 0.68 & 0.67 & 0.67 & 0.54 & 0.69 \\ \hline
        Llama-2 13B Chat & 0.34 & 0.37 & 0.41 & 0.33 & 0.44 & 0.27 & 0.38 \\ \hline
        Llama-2 7B Chat & 0.38 & 0.32 & 0.42 & 0.44 & 0.46 & 0.28 & 0.38 \\ \hline
        Llama-2 7B & 0.29 & 0.34 & 0.28 & 0.34 & 0.19 & 0.31 & ~ \\ \hline
        Phi-2 & 0.26 & 0.21 & 0.26 & 0.29 & 0.24 & 0.3 & 0.26 \\ \hline
        \textbf{LLM/Prompting Scheme} & \textbf{Task-Only} & \textbf{Task Definition} & \textbf{Context Analyze} & \textbf{Context Question} & \textbf{Few Shot}  & \textbf{Zero-shot CoT} & \textbf{CODA} \\ \hline
        \textbf{srq} & ~ & ~ & ~ & ~ & ~ & ~ & ~ \\ \hline
        Baseline & 0.45 & ~ & ~ & ~ & ~ & ~ & ~ \\ \hline
        Falcon 40B Instruct & 0.31 & 0.31 & 0.39 & 0.39 & 0.37 & 0.41 & ~ \\ \hline
        Falcon 7B Instruct & 0.22 & 0.26 & 0.36 & 0.39 & 0.41 & 0.36 & 0.28 \\ \hline
        T5 3B Flan-Alpaca & 0.3 & 0.29 & 0.41 & 0.38 & 0.48 & 0.43 & 0.3 \\ \hline
        T5 11B Flan-Alpaca & 0.31 & 0.3 & 0.41 & 0.41 & 0.47 & 0.47 & 0.36 \\ \hline
        Mistral v0.1 7B Instruct & 0.33 & 0.34 & 0.3 & 0.37 & 0.42 & 0.45 & 0.3 \\ \hline
        UL2 20B Flan & 0.29 & 0.32 & 0.47 & 0.51 & \textbf{0.56} & 0.5 & 0.33 \\ \hline
        Llama-2 13B Chat & 0.33 & 0.31 & 0.4 & 0.4 & 0.48 & 0.38 & 0.32 \\ \hline
        Llama-2 7B Chat & 0.32 & 0.32 & 0.37 & 0.4 & 0.45 & 0.3 & 0.33 \\ \hline
        Llama-2 7B & 0.27 & 0.31 & 0.33 & 0.36 & 0.33 & 0.33 & ~ \\ \hline
        Phi-2 & 0.24 & 0.27 & 0.31 & 0.34 & 0.28 & 0.35 & 0.25 \\ \hline
        \textbf{LLM/Prompting Scheme} & \textbf{Task-Only} & \textbf{Task Definition} & \textbf{Context Analyze} & \textbf{Context Question} & \textbf{Few Shot}  & \textbf{Zero-shot CoT} & \textbf{CODA} \\ \hline
        \textbf{wtwt} & ~ & ~ & ~ & ~ & ~ & ~ & ~ \\ \hline
        Baseline & 0.62 & ~ & ~ & ~ & ~ & ~ & ~ \\ \hline
        Falcon 40B Instruct & 0.53 & \textbf{0.56} & 0.53 & 0.54 & 0.4 & 0.38 & ~ \\ \hline
        Falcon 7B Instruct & 0.31 & 0.32 & 0.15 & 0.14 & 0.35 & 0.17 & 0.38 \\ \hline
        T5 3B Flan-Alpaca & 0.41 & 0.40 & 0.4 & 0.42 & 0.43 & 0.54 & 0.41 \\ \hline
        T5 11B Flan-Alpaca & 0.39 & 0.36 & 0.53 & 0.52 & \textbf{0.58} & 0.55 & 0.53 \\ \hline
        Mistral v0.1 7B Instruct & 0.41 & 0.41 & 0.52 & 0.49 & 0.56 & 0.34 & 0.33 \\ \hline
        UL2 20B Flan & 0.39 & 0.37 & 0.44 & 0.43 & 0.57 & 0.48 & 0.43 \\ \hline
        Llama-2 13B Chat & 0.4 & 0.38 & 0.47 & 0.45 & 0.28 & 0.29 & 0.41 \\ \hline
        Llama-2 7B Chat & 0.38 & 0.38 & 0.28 & 0.15 & 0.47 & 0.27 & 0.32 \\ \hline
        Llama-2 7B & 0.32 & 0.34 & 0.29 & 0.27 & 0.35 & 0.3 & ~ \\ \hline
        Phi-2 & 0.33 & 0.32 & 0.37 & 0.29 & 0.39 & 0.37 & 0.32 \\ \hline
        \caption{Average unweighted F1-scores. The benchmark results are given in the first row from the original paper. The highest scoring result for each data set is in \textbf{bold}. Note that we were unable to run the CoDA architecture on the larger prompts.}
        \label{tab:results}
\end{longtable}

\subsection{Results of Fine-Tuning LLMs}
\autoref{tab:fine_tune_results} shows the results of fine-tuning selected LLMs on the datasets. In general, fine-tuning does not perform better than zero-shot LLMs for out-of-domain data. In fact, in most cases, LLMs performed worse compared to the zero-shot LLM setup and even the baselines. For example, the fine-tuned models tended to be much worse than zero-shot models on SemEval2016 and election2016, but better on covid-lies and phemerumours. This result may be due to how stance was operationalized differently between different datasets \cite{ng2022my}. It could also be due to the fact that fine-tuning could decrease the generalizability of the LLMs, as observed in past work, which includes fine-tuned LLMs for domains such as code completion \cite{yuan2023evaluating,zhang2023scaling,kumar2022fine}. 

\begin{table*}[!ht]
\centering
\begin{tabular}{|l|l|l|l|l|l|l|}
\hline
\textbf{Dataset/Model} & covid-lies & election2016 & phemerumors & SemEval2016 & srq & wtwt \\ \hline 
T5 11B Flan-Alpaca & 0.40 & 0.29 & 0.35 & 0.19 & 0.33 & 0.32 \\ \hline
UL2 20B Flan & 0.27 & 0.43 & 0.21 & 0.47 & 0.42 & 0.40 \\ \hline 
Mistral v0.1 7B Instruct & 0.41 & 0.46 & 0.45 & 0.45 & 0.18 & 0.53 \\ \hline
Llama-2 7B Chat & 0.34 & 0.33 & 0.34 & 0.29 & 0.35 & 0.32 \\ \hline 
\end{tabular}
\caption{Summary of Results of Fine-Tuning selected LLMs. The dataset tested on is the out-of-domain data, while the models were fine-tuned with all other data.}
\label{tab:fine_tune_results}
\end{table*}

\subsection{Evaluation of Valid Responses from Models }
We find that there are a number of completions in which the LLM does not give a valid stance label in response to a prompt. With this, we find that once again the encoder-decoder architectures and FSP and CoT prompting schemes tend to produce more valid completions and are more accurate when they do produce a valid completion. Along with that, even though all prompts explicitly tell the LLM to output a single label, many completions have longer text than just a single label. 

\autoref{tab:goodresults_f1} shows the accuracy scores of the good results derived from the zero-shot LLMs. \autoref{tab:fine_tune_goodoutputs} shows the accuracy scores of the good results from the fine-tuned LLMs. We observe that the accuracy scores for the good results consistently and significantly outperform the baseline supervised approaches. However, when there is ambiguity in the response, the model performance drops drastically. This shows that the output response is an indicator of model performance in the task of stance detection on any given example. We observe that FSP and CoT have the best overall scores, suggesting the superiority of these prompting methods.

Furthermore, we analyzed the response lengths of the models. Despite being prompted to only return a single stance label, many models returned additional words. \autoref{tab:text_length} presents the average response length of the models. The encoder-decoder family of models faithfully follow this instruction by only outputting a single word for stance. However, Llama-2 models tend to produce longer responses. Doing a correlation analysis, we find that having a longer text output weakly correlates to worse stance predictions. For example, we find that when we look at good labels, the correlation between the length of the output and whether the output is the correct stance label is $-0.02$ ($p<0.001$). With these observations, we constructed a simple decision tree model using the Python sklearn package to test whether we can predict if the LLM had predicted the stance correctly. Using the raw output length, number of non-stance words in the output and whether the output contains a valid stance label as independent variables, we can predict with 0.59 accuracy whether the LLM predicted the stance correctly, regardless of what input it was fed. In other words, there is a distinct relation between the quality of the output in terms of producing a valid stance label and following the output instructions, and whether the LLM was correct in predicting the stance.

\begin{longtable}{|l|l|l|p{1.5cm}|p{1.5cm}|l|p{1.5cm}|l|}
    \hline
        \textbf{LLM/Prompting Scheme} & \textbf{Task-Only} & \textbf{Task Definition} & \textbf{Context Analyze} & \textbf{Context Question} & \textbf{Few Shot}  & \textbf{Zero-shot CoT} & \textbf{CODA} \\ \hline
        Llama-2 13B Chat & 0.83 & 0.81 & 0.57 & 0.70 & 0.99 & 0.84 & 0.74 \\ \hline 
        Llama-2 7B Chat & 0.90 & 0.72 & 0.89 & 0.70 & 0.99 & 0.91 & 0.72 \\ \hline 
        Llama-2 7B & 0.40 & 0.56 & 0.76 & 0.57 & 0.33 & 0.68 & ~ \\ \hline 
        Mistral v0.1 7B Instruct & 0.99 & 0.99 & 0.99 & 0.99 & 0.99 & 0.99 & 0.81 \\ \hline 
        Falcon 40B Instruct & 0.99 & 0.99 & 0.99 & 0.99 & 0.99  & 0.84 & ~ \\ \hline 
        Falcon 7B Instruct & 0.96 & 0.94 & 0.95 & 0.96 & 0.88 & 0.96 & 0.84 \\ \hline 
        T5 3B Flan-Alpaca & 0.99 & 0.99 & 0.99 & 0.99 & 0.99 & 0.99 & 0.99 \\ \hline 
        T5 11B Flan-Alpaca & 0.99 & 0.99 & 0.99 & 100 & 100 & 100 & 100 \\ \hline 
        UL2 20B Flan & 100 & 100 & 100 & 100 & 100 & 100 & 0.99 \\ \hline 
        Phi-2 & 0.74 & 0.61 & 0.87 & 0.71 & 0.58 & 0.67 & 0.37 \\ \hline 
    \caption{F1 score of accuracy of `good outputs' of zero-shot LLMs}
    \label{tab:goodresults_f1}
\end{longtable}

\begin{table*}[!ht]
\centering
\begin{tabular}{|l|l|}
\hline
\textbf{Model} & \textbf{Performance} \\ \hline 
Llama-2 7B & 67.36 \\ \hline 
Mistral v0.1 7B Instruct & 99.99 \\ \hline 
T5 11B Flan-Alpaca & 94.49 \\ \hline 
UL2 20B Flan & 99.89 \\ \hline 
\end{tabular}
\caption{F1 score of accuracy of the `good outputs' of the fine-tuned models}
\label{tab:fine_tune_goodoutputs}
\end{table*}

\begin{table*}[!ht]
\centering
\begin{tabular}{|l|l|}
\hline
\textbf{Model} & \textbf{Avg Response Length} \\ \hline 
Falcon 40B Instruct & 9.59 \\ \hline 
Falcon 7B Instruct & 1.60 \\ \hline 
T5 3B Flan-Alpaca & 1.0 \\ \hline 
T5 11B Flan-Alpaca & 1.0 \\ \hline 
Llama-2 13B Chat & 11.65 \\ \hline 
Llama-2 7B Chat & 7.40 \\ \hline 
Llama-2 7B & 20.97 \\ \hline 
Mistral v0.1 7B Instruct & 2.71 \\ \hline
UL2 20B Flan & 1.0 \\ \hline 
Phi-2 & 22.05 \\ \hline 
\end{tabular}
\caption{Average response length of models by the number of words the model outputs for a stance label}
\label{tab:text_length}
\end{table*}

\section{Discussion}
\label{sec:discussion}
In this section we discuss the findings of the various tests we used to investigate the use of LLMs for the task of stance classification.

\paragraph{\textbf{LLMs are Competitive with In-Domain Supervised Models but have Inconsistent Performance}}
We evaluated open-source LLMs because these are the most cost-effective LLMs and thus are likely to have higher usage by analysts. Furthermore, previous research has highlighted that there is likely contamination of data with GPT models that could skew the results \cite{aiyappa2023can}. In our analysis, we observe that LLMs do not consistently outperform their baseline cousins. We do find that encoder-decoder LLM architectures tend to perform better as compared to the decoder-only architectures. We also find that a larger model size does not necessarily lead to an improvement in performance, hence we suggest that while using LLMs for stance detection, an appropriately chosen smaller-sized LLM could suffice, which can alleviate the computational burden for inference. Furthermore, given the inconsistent performance of LLMs in the task, we also recommend following something like a Data Programming paradigm \cite{ratner2016data} to use LLM generated labels in combination with other signals of stance to classify the stance of statements.

During our testing, we also found that there are a number of outputs where an LLM does not give a valid stance label in response to a prompt. We find that encoder-decoder based architectures together with FSP and CoT prompting schemes tend to produce more valid results. This reflects the instability of LLMs, in which they do not always produce the desired output even though we had specified the output format. This observation could also be a result of using a completion format as a prompt, in which the model takes it as a paragraph or sentence completion task after processing the information given prior in the prompt. Where there is a valid completion, the models perform generally well, and often better than in-domain supervised models. However, performance drops when there is ambiguity in the output; therefore, indicating the formatting of the output response is crucial to achieving performance with LLMs. Downstream work can facilitate the crafting of prompt templates or the tweaking of LLMs architecture to produce consistent and valid outputs.

\paragraph{\textbf{Importance of Prompting Scheme Development}} The testing of an array of prompting schemes reveals the importance of the input prompt into the LLMs. The performance of any given LLM varies depending on the prompt format, suggesting that the prompting scheme and its formatting are crucial to eliciting good performance from the LLM. In general, the use of FSP and CoT prompting improves performance of the model, however, prompting schemes can also be model dependent. There are instances that the prompt does improve performance of the LLM over the baselines which use traditional machine learning methods. The baseline methods do not account for the target the sentence is directed towards, suggesting that the insertion of target and additional contextual information improves the performance.

\paragraph{\textbf{Fine-Tuning Does not Necessarily Lead to Better Performance}}
Our results find that fine-tuning the LLMs to create a specialized model does not lead to better performance of the models. This is likely because the fine-tuning makes the model too specialized, and thus unable to generalize to out-of-domain data points and varying defintions of stance. Previous studies also discover that fine-tuning does not work well when the size of the model exceeds the size of the fine-tuning data, especially when working with out-of-domain data \cite{zhang2023scaling,kumar2022fine}. In our case, we only have a few thousand data points against model sizes in billions. Therefore, a fine-tuned LLM is not as desirable as a zero-shot LLM, which can generalize better.

\paragraph{\textbf{Stance detection as an LLM Benchmark}}
LLMs are built and assessed on a range of benchmarks, from common sense reasoning, reading comprehension, to mathematical reasoning \cite{zheng2023judging}. However, one language task not represented in current LLM benchmarking is the task of stance classification. This task is likely overlooked because it is not a task as part of the standard natural language processing evaluation playbook. The task of stance classification is, however, a crucial benchmark to investigate, as many policy formulations depend on understanding public opinion, for example \cite{upadhyaya2023multi}. Misclassification of sentence stances can lead to incorrect interpretations of opinion slant. Stances, especially stances in social media posts towards a target, are used in downstream analyses that evaluate aggregated stances towards topics to understand public reaction and formulate policy. In those scenarios, incorrect stance classifications can result in erroneous interpretations and policy mismatches \cite{alturayeif2023systematic}.

Given the importance of stance classification to the wider society and the results of this study, we would like to propose that stance classification be considered as a future benchmarking task for LLMs. 

\paragraph{\textbf{Limitations}}
As in all studies, several limitations nuance our work. For our stance detection task, our data sets are premised on manual annotations, which could be subjected to inconsistent annotations and varied sentence interpretations \cite{ng2022my}. For example, there is a sentence about Michael Essien having Ebola, ``@xx no he hasn't. The man himself confirmed not true @MichaelEssien" that was annotated as a neutral stance whereas it should be a stance \textit{against} the claim that Michael Essien had contracted Ebola.

Next, in all research involving LLMs, a primary concern arises from the datasets used to pre-train LLMs, from which they acquire their language capabilities and knowledge base. These datasets may harbor inherent biases or offensive content \cite{dataconcerns}. Although this study makes no attempt to exploit or introduce any form of bias, it is possible that these biases might inadvertently permeate the analysis performed using LLMs, especially when performing politically-charged tasks like stance detection. This underlines the importance of diligently scrutinizing the data used to train LLMs, especially when they are employed in socially significant tasks such as stance classification, where bias can have profound implications.

Additionally, while we attempted to consider a wide range of open-source models and possible configurations for those models, we were constrained by computational resources and time from using every possible permutation of open-source models for the task of stance classification. We believe that we have tested a representative sample of offerings, however, it is possible that a certain LLM, perhaps due to pre-training data or even architectural differences, may actually perform uncharacteristically in regards to the models tested in this study. Additionally, we also only used greedy decoding for the LLMs and manual selection of examples for the FSP prompts. Its possible that different decoding techniques or a more optimal example selection for FSP (e.g., \cite{lu2021fantastically}) could have produced better results. 

\paragraph{\textbf{Ethical Considerations}}

Lastly, using LLMs for stance detection at scale presents an ethical consideration that pertains to the environmental impact of running these computationally intensive models. It is indisputable that LLMs consume more energy compared to their smaller counterparts. Thus, their use in tasks like stance classification is associated with a tangible energy cost. However, it is also crucial to balance this against the alternative scenario, which involves continuous human effort for labeling data, a process that is both labor-intensive and time-consuming due to the generalizability problem inherent in stance classification.

Looking ahead, as we strive to create more sustainable and efficient language models, one potential avenue could be leveraging LLMs to distill smaller, more energy-efficient models for production purposes. This could significantly decrease the energy demand, making stance classification tasks more environmentally sustainable, while still benefiting from the superior performance of LLMs. As our understanding of LLMs continues to evolve, it is paramount to remain vigilant about these ethical considerations and strive towards more responsible and sustainable practices.

Additionally, as with any classification effort of text, efforts towards stance detection could be used for text-based censorship. For example, it is possible that our research could be used to identify, at scale, comments and users that a presenting a certain stance toward a target and then moderate those users or comments, thus maintaining a harmonious digital society. While there can be misuse towards using stance detection models to identify and retain toxic comments rather than remove them, we believe that the benefits of being able to more correctly classify stances of text comments outweigh its potential misuse, and that the same precautions used to prevent misuse with the classification of texts more broadly can also be applied to this work.

\section{Concluding Remarks}
\label{sec:conclusion}
Stance classification is a tricky problem. It requires natural language understanding to sieve out the opinion of the author towards a target, and involves many language nuances and subtleties, and is context dependent. Through this paper, we shed light on the stance classification abilities of LLMs, contributing valuable insights that can guide future advancements in this domains. From a series of extensive experiments with stance detection using a large series of Large Language Models, we observed the behavior and performance of LLMs on classifying the stance of social media datasets. We summarize our observations as follows:
\begin{enumerate}
    \item Large Language Models do not consistently outperform supervised machine learning models for the stance classification task. They are sensitive to the input instructions, therefore the formulation of a prompting scheme for input is crucial.
    \item Performance accuracy for stance classification varies considerably across models and prompts. At the current stage of LLM development, we suggest to use encoder-decoder based model architectures together with FSP or CoT prompting schemes for best performance.
    \item Fine-tuning LLMs does not necessarily improve the out-of-domain performance of the model, and can in fact worsen it. Thus, a more generalized model is preferred over a specialized one.
\end{enumerate}

The application of LLMs on the stance detection task not only streamlines the process but also paves the way for expanding stance detection capabilities across languages and domains. Given our observation that LLMs do sometimes perform better than the baseline supervised models, there is hope of improvement of stance classification capabilities of LLMs. As stance detection is a crucial task in many workstreams that analyze the opinion of the public, we urge the world of natural language processing to join forces and build a set of practices for the use of LLMs in analyzing social data like in this stance detection problem. Such sets of practices should reduce inconsistencies between and within models, improving the usage of the models in downstream tasks. 

\begin{acks}
The research for this paper was supported in part by the Center for Informed Democracy and Social-cybersecurity (IDeaS) and the Center for Computational Analysis of Social and Organizational Systems (CASOS) at Carnegie Mellon University. This work was also conducted within the Cognitive Security Research Lab at the Army Cyber Institute at West Point and supported in part by the Office of Naval Research (ONR) under Support Agreement No. USMA 20057. The views and conclusions are those of the authors and should not be interpreted as representing the official policies, either expressed or implied, of the Department of Defense, the U.S. Army, or the U.S. Government.
\end{acks}

\bibliographystyle{ACM-Reference-Format}
\bibliography{main}

\appendix
\section{Examples of Prompting Schemes}
\label{sec:prompt_examples}
\autoref{tab:prompt_examples} provides examples of the prompts constructed as inputs to the LLMs.

\begin{longtable}{|p{4cm}|p{8cm}|}
        \hline
        Task-only & ```Analyze the following statement and determine its stance. Respond with a single word: "for", "against", "neutral", or "unrelated". Only return the stance as a single word, and no other text.
        
        statement: \{statement\}
    
        stance:'''\\ \hline 
        Task Definition & ```Stance classification is the task of determining the expressed or implied opinion, or stance, of a statement toward a certain, specified target. Analyze the following statement and determine its stance. Respond with a single word: "for", "against", "neutral", or "unrelated". Only return the stance as a single word, and no other text.
        
        statement: \{statement\}
    
        stance:'''\\ \hline 
        Context Analyze & ```Stance classification is the task of determining the expressed or implied opinion, or stance, of a statement toward a certain, specified target. Analyze the following social media statement and determine its stance towards the provided entity. Respond with a single word: "for", "against", "neutral", or "unrelated". Only return the stance as a single word, and no other text.

        entity: \{event\}
        
        statement: \{statement\}
    
        stance:'''\\ \hline 
        Context Question & ```Stance classification is the task of determining the expressed or implied opinion, or stance, of a statement toward a certain, specified target. What is the stance of the following social media statement toward the following entity? Respond with a single word: "for", "against", "neutral", or "unrelated". Only return the stance as a single word, and no other text.

        entity: \{event\}
        
        statement: \{statement\}
    
        stance:'''\\ \hline 
        Few-shot Prompt (FSP) & ```Stance classification is the task of determining the expressed or implied opinion, or stance, of a statement toward a certain, specified target. The following statements are social media posts expressing opinions about entities. Each statement can either be for, against, neutral, or unrelated toward their associated entity.
        
        entity: Atheism
        
        statement: Leaving Christianity enables you to love the people you once rejected. \#freethinker \#Christianity \#SemST
        
        stance: for
        
        entity: Climate Change is a Real Concern
        
        statement: \@AlharbiF I'll bomb anything I can get my hands on, especially if THEY aren't christian. \#graham2016 \#GOP \#SemST
        
        stance: neutral
        
        entity: Feminist Movement
        
        statement: Always a delight to see chest-drumming alpha males hiss and scuttle backwards up the wall when a feminist enters the room. \#manly \#SemST
        
        stance: for
        
        entity: Hillary Clinton
        
        statement: Would you wanna be in a long term relationship with some bitch that hides her emails, \& lies to your face? Then \#Dontvote \#SemST
        
        stance: against
        
        entity: Legalization of Abortion
        
        statement: \@k\_yoder That lady needs help, mental illness is a serious issue. \#SemST
        
        stance: neutral
        
        Analyze the following social media statement and determine its stance towards the provided entity. Respond with a single word: "for", "against", "neutral", or "unrelated". Only return the stance as a single word, and no other text.
        
        entity: \{event\}
        
        statement: \{statement\}
        
        stance:'''\\ \hline
        Zero-shot Chain-of-Thought (CoT) & ```Stance classification is the task of determining the expressed or implied opinion, or stance, of a statement toward a certain, specified target. Think step-by-step and explain the stance (for, against, neutral, or unrelated) of the social media statement towards the entity.
        
        entity: \{event\}
        
        statement: \{statement\}
        
        explanation:'''
        
        \textit{LLM Replies}
        
        ```Therefore, based on your explanation, \{stance\_reason\}, what is the final stance? Respond with a single word: "supports", "denies", "neutral", or "unrelated". Only return the stance as a single word, and no other text.
        
        entity: \{event\}
        
        statement: \{statement\}
        
        stance:'''\\ \hline
        CoDA Prompt &  ```Accurately and concisely explain the linguistic elements in the statment and how these elements affect meaning, including grammatical structure, tense and inflection, virtual speech, rhetorical devices, lexical choices and so on. Do nothing else.
        
        statement: \{statement\}
        
        explanation:'''

        \textit{LLM gives linguist response}

        ```Accurately and concisely explain the key elements contained in the following statement, such as characters, events, parties, religions, etc. Also explain their relationship with \{event\}. Do nothing else.
        
        statement: \{statement\}
        
        explanation:'''

        \textit{LLM gives expert response}

        ```Analyze the following statement, focusing on the content, hashtags, Internet slang and colloquialisms, emotional tone, implied meaning, and so on. Do nothing else.
        
        statement: \{statement\}
        
        explanation:'''

        \textit{LLM gives social media user response}

        ```You think the attitude behind the following statement is in support of \{event\}. Following the statement are explanations of the statement by various personas. Identify the top three pieces of evidence from these that best support your opinion and argue for your opinion.
        
        statement: \{statement\}
        
        linguist explanation: \{linguist\_analysis\}
        
        expert explanation: \{expert\_analysis\}
        
        heavy social media user explanation: \{user\_analysis\}
        
        opinion:'''

        \textit{LLM gives `for' argument}

        ```You think the attitude behind the following statement is against \{event\}. Folloiwng the statement are explanations of the statement by various personas. Identify the top three pieces of evidence from these that best support your opinion and argue for your opinion.
        
        statement: \{statement\}
        
        linguist explanation: \{linguist\_analysis\}
        
        expert explanation: \{expert\_analysis\}
        
        heavy social media user explanation: \{user\_analysis\}
        
        opinion:'''

        \textit{LLM gives `against' argument}

        ```Determine whether the following statement is in favor of, neutral, against, or unrelated to \{event\}. 
        
        statement: \{statement\}
        
        Arguments that the attitude is in favor: \{for\_opinion\}
        
        Arguments that the attitude is against: \{against\_opinion\}
        
        Choose the stance from "for", "against", "neutral", or "unrelated". Answer with only the option above that is most accurate and nothing else.
        stance:''' \\ \hline
    \caption{Examples of prompts constructed for each prompting scheme. These examples are constructed for the SemEval2016 dataset.}
    \label{tab:prompt_examples}
\end{longtable}

\section{Graphical Results}
\label{sec:graphical_results}
\begin{figure}  
    \centering
\includegraphics[scale=0.4]{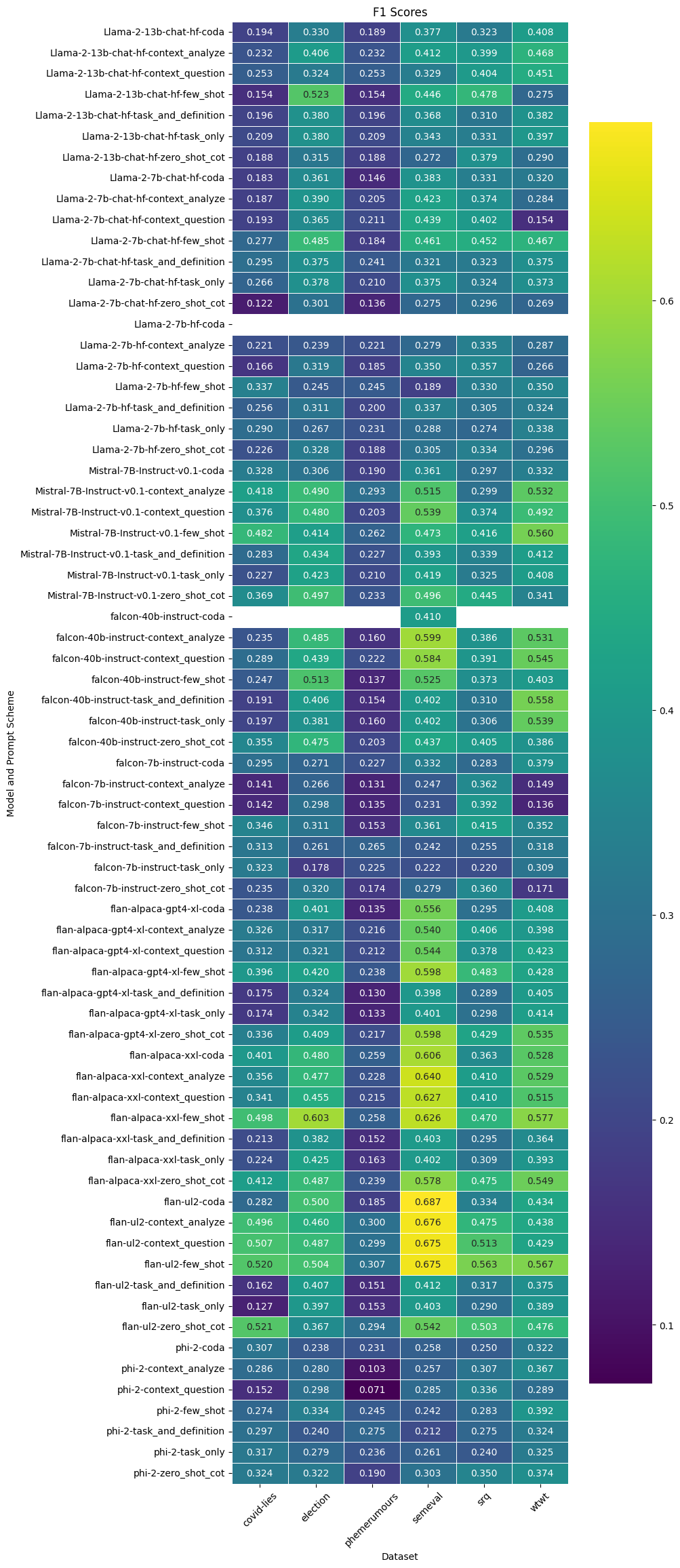}
    \caption{Graphical Representation of all results by F1 score accuracy}
    \label{fig:all_graphical}
\end{figure}

\begin{figure}  
    \centering
\includegraphics[width=\linewidth]{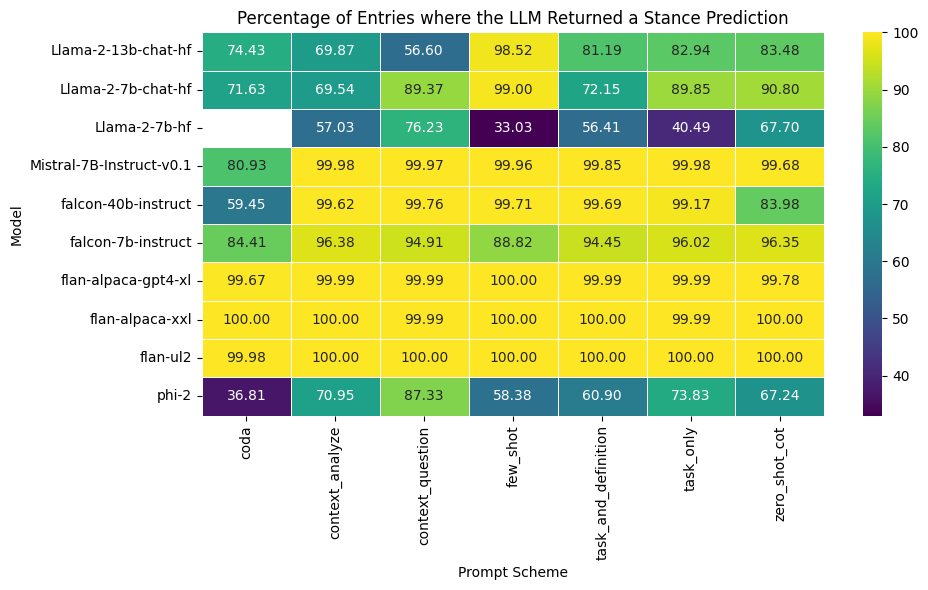}
    \caption{Graphical presentation of the proportion of ooutputs that returns a stance prediction}
    \label{fig:return_prediction}
\end{figure}

\begin{figure}  
    \centering
\includegraphics[width=\linewidth]{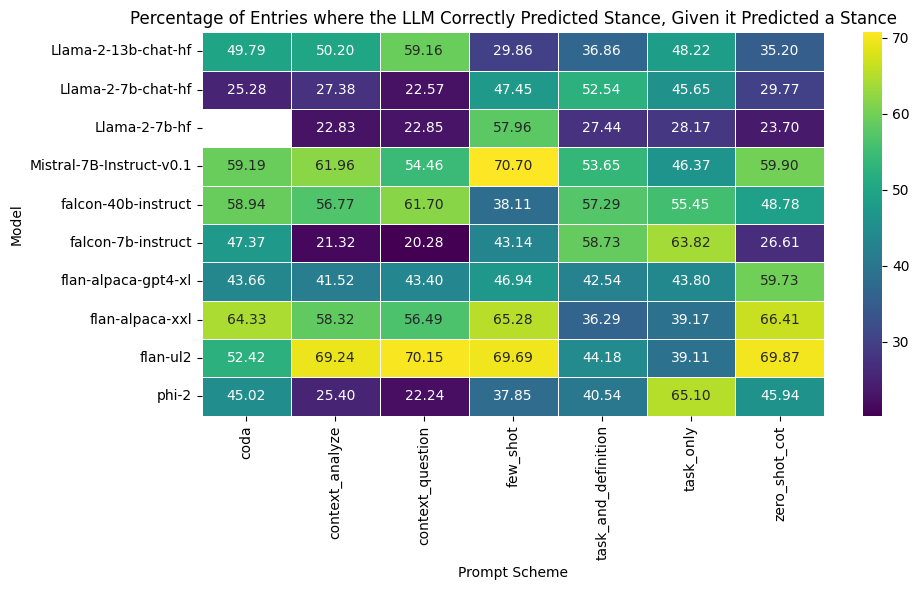}
    \caption{Proportion of entries that the LLM correctly predicted the stance, given that it had a valid stance.}
    \label{fig:correct_valid}
\end{figure}

This section presents a graphical view of the results in terms of heatmap. \autoref{fig:all_graphical} presents the accuracy scores of all the results. \autoref{fig:return_prediction} presents the proportion of outputs that returns a valid stance prediction. \autoref{fig:correct_valid} presents the scores where the LLM correctly predicted the stance, given it had a valid stance.

\end{document}